\documentclass{article}

 \usepackage[dblblindworkshop, final]{neurips_2026}
\workshoptitle{I Can't Believe It's Not Better: Failure Modes of AI in Biology}

\usepackage[utf8]{inputenc} 
\usepackage[T1]{fontenc}    
\usepackage{hyperref}       
\usepackage{url}            
\usepackage{booktabs}       
\usepackage{amsfonts}       
\usepackage{nicefrac}       
\usepackage{microtype}      
\usepackage{xcolor}         

\usepackage{amsmath}
\usepackage{algorithm}
\usepackage{algpseudocode}
\usepackage{graphicx}
\usepackage{colortbl}
\usepackage{multirow} 
\usepackage{amssymb}
\usepackage{subcaption}

\title{When Genomic Masking Priors Fail to Transfer: Strong Variant Prediction, Weak Functional Generation}
\workshoptitle{I Can't Believe It's Not Better: Failure Modes of AI in Biology}

\author{%
  \textbf{Susu Hu}\textsuperscript{1,2,3,4}%
  \thanks{Corresponding author: susu.hu@outlook.com 
  \textsuperscript{1}Translational Surgical Oncology, National Center for Tumor Diseases (NCT/UCC), Dresden, Germany 
  \textsuperscript{2}Faculty of Medicine and University Hospital Carl Gustav Carus, TU Dresden, Germany \newline
  \textsuperscript{3}Helmholtz-Zentrum Dresden-Rossendorf (HZDR), Dresden, Germany 
  \textsuperscript{4}German Cancer Research Center (DKFZ), Heidelberg, Germany 
  \textsuperscript{5}ScaDS.AI -- Dresden University of Technology 
  \textsuperscript{6}TUD Dresden University of Technology, Germany 
  \textsuperscript{7}Amazon (work done outside of Amazon), Dresden, Germany 
  \textsuperscript{8}Leibniz Information Centre for Science and Technology, Hannover, Germany
  \textsuperscript{9}Drug Development Department, Gustave Roussy, Villejuif, France 
  \textsuperscript{10}Université Paris-Saclay, Gustave Roussy, Inserm U981, IHU PRISM, Villejuif, France 
  \textsuperscript{11}Medical Oncology, National Center for Tumor Diseases (NCT), University Hospital Heidelberg, Heidelberg, Germany%
  } \\
  \And
  \textbf{Preetam Gattogi}\textsuperscript{5} \\
  \And
  \textbf{Jens Lehmann}\textsuperscript{6,7} \\
  \And
  \textbf{Sahar Vahdati}\textsuperscript{5,8} \\
  \And
  \textbf{Stefanie Speidel}\textsuperscript{1,2,3,4} \\
  \And
  \textbf{Julien Vibert}\textsuperscript{9,10,11}
}

\begin{document}

\maketitle

\begin{abstract}
Bidirectional discrete diffusion model appears naturally suited
to genomic modeling because it can reconstruct missing sequence from both
flanks. We developed \textbf{GenDA} (\textbf{Gen}omic
\textbf{D}ensity-optimized \textbf{A}bsorbing Diffusion) under the additional
hypothesis that entropy-guided span placement would concentrate reconstruction
pressure on compositionally complex regions, improving both downstream
variant-effect prediction and functional sequence generation. Our results only
partially support this premise. After supervised fine-tuning, the
202M-parameter GenDA model reaches a pooled ClinVar SNV AUROC of
0.774, exceeding a similarly scaled autoregressive model by 0.103.
However, a matched random-span variant reaches 0.777, providing no evidence
that entropy guidance causes the ClinVar improvement. More unexpectedly, GenDA
fails a zero-shot functional inpainting stress test: across promoters,
enhancers, exon boundaries, and intron boundaries, it does not consistently
outperform a control that shuffles the native gap while exactly preserving
3-mer composition. Failure is already present for 50--500-bp gaps, although
enhancer degradation worsens at longer gaps. Diagnostics identify several
boundary conditions: entropy measures local sequence complexity rather than
functional importance; 1-mer tokenization limits physical context; training
spans are capped at 300 bp; and high absolute AlphaGenome fidelity can coexist
with negative control-normalized restoration. These results show that strong
fine-tuned variant prediction, a plausible corruption prior, and functional
generation are distinct claims that require separate validation. \textit{Code is available at: https://github.com/susuhu/GenDA}
\end{abstract}

\section{Introduction}

Progress on genomic foundation models is usually established through held-out
likelihood, variant prediction, or supervised downstream benchmarks. These
measurements are valuable, but they do not establish that a model can generate
sequence that preserves locus-specific molecular function. The distinction is
especially important for genomic inpainting: many nucleotide sequences can be
statistically plausible, yet only a small subset may maintain promoter output,
enhancer activity, or splice usage in the original context.

Bidirectional masked diffusion appears well suited to genomic inpainting.
Unlike a left-to-right autoregressive generator, it can condition directly on
both flanks of a missing region and revise uncertain positions iteratively.
Contiguous span corruption should further encourage reconstruction from
broader context rather than from immediately adjacent visible nucleotides.

GenDA was designed around a further, less certain premise. We hypothesized that uniform  masking spends substantial training capacity on locally simple or repetitive sequence, whereas entropy-guided span placement would more frequently
mask compositionally complex regions that are harder to reconstruct. If this
prior induced more informative reconstruction pressure, we expected its
benefits to appear in both fine-tuned variant-effect prediction and zero-shot
functional inpainting. Local entropy was not assumed to identify function
directly; the hypothesis was that reconstruction difficulty would provide a
useful label-free training signal.

We test these expectations using \textbf{GenDA} (\textbf{Gen}omic
\textbf{D}ensity-optimized \textbf{A}bsorbing Diffusion), a 202M-parameter
model trained on heterogeneous 4,096-bp human genomic windows. After supervised
fine-tuning, GenDA reaches 0.774 pooled ClinVar SNV AUROC, compared with 0.671
for a similarly scaled AR baseline. However, Random Span reaches 0.777, so the
discriminative result cannot be attributed to entropy guidance. More
unexpectedly, GenDA, AR Llama, Evo~2, and D3LM all fail to consistently beat an
exact 3-mer-preserving control during zero-shot functional inpainting.
Promoter and enhancer restoration remains negative even for 50--500-bp gaps.
The paper therefore examines both a failed masking prior and a failure of
strong fine-tuned prediction to identify generative utility.

This paper makes three contributions:
\begin{enumerate}
    \item We document a discrimination--generation disconnect: a
    model that improves pooled ClinVar AUROC by 0.103 over a similarly scaled
    AR baseline does not preserve AlphaGenome-predicted function during
    zero-shot inpainting.
    \item We identify plausible boundary conditions using corruption-geometry,
    masking, metric, and decoding diagnostics. The model covers only 4,096 bp,
    and its 300-bp training-span cap is substantially shorter than the largest
    3,500-bp evaluation gap.
    \item We derive actionable evaluation recommendations: report
    composition-preserving controls, distinguish absolute fidelity from
    control-normalized restoration, stratify by trained corruption scale, and
    evaluate discriminative and generative capabilities separately.
\end{enumerate}

\section{Related Work}

\paragraph{Genomic representation and generation.}
Masked models such as DNABERT-2~\citep{zhou2023dnabert} and the Nucleotide
Transformer~\citep{dalla2025nucleotide} learn useful representations, while
HyenaDNA~\citep{nguyen2023hyenadna}, Evo~\citep{nguyen2024sequence}, and
Evo~2~\citep{brixi2025genome} emphasize scalable AR generation. Caduceus
incorporates reverse-complement equivariance for representation learning
\citep{schiff2024caduceus}. These lines of work establish that architecture and
pretraining can improve biological prediction, but prediction quality and
conditional generation are generally evaluated separately.

\paragraph{Discrete genomic diffusion and comparison scope.}
Discrete diffusion frameworks include D3PM~\citep{austin2021structured},
MaskGIT~\citep{chang2022maskgit}, MDLM~\citep{sahoo2024simple}, and
absorbing-state models such as LLaDA~\citep{nie2025large}. Genomic adaptations
include D3~\citep{sarkar2024designing} and D3LM~\citep{yang2026d3lm}; continuous
or simplex alternatives include DNA-Diffusion~\citep{dasilva2024dna} and
Dirichlet flow matching~\citep{stark2024dirichlet}. Reward and constraint
methods can improve selected properties after training
\citep{wang2024fine,chen2025ctrldnacontrollablecelltypespecificregulatory,uehara2025reward}.
These are not interchangeable baselines: the first group provides algorithms,
not pretrained genomic checkpoints, while most genomic design systems target
short, curated regulatory sequences using specialized conditioning, continuous
relaxations, or rewards. Scaling them to heterogeneous 4,096-bp sequences at
single-nucleotide resolution remains unestablished, and retraining them would
introduce new tuning choices rather than a controlled substitution. We use the
matched causal model and masking ablations for controlled comparisons, with
D3LM and Evo~2 as descriptive pretrained references. We ask what evidence is
required before a diffusion model is considered useful for functional
generation.

\paragraph{Predictor-based evaluation.}
Sequence-to-function models enable scalable evaluation of generated DNA, but
their predictions remain proxies rather than experimental measurements. We use
AlphaGenome because it produces locus-resolved CAGE, chromatin, and splice
predictions from long genomic context~\citep{avsec2026advancing}. Crucially, we
compare generated gaps with composition-preserving corruptions; otherwise high
agreement with the native prediction may reflect unchanged context, smooth
outputs, or low assay sensitivity rather than successful reconstruction.

\section{Design Premise}

\paragraph{Hypotheses under test.}
GenDA tests three linked expectations: (i) contiguous spans provide a better
reconstruction signal than exclusive independent 1-mer masking; (ii)
entropy-guided starts improve over random spans by emphasizing compositionally
complex sequence; and (iii) the resulting pretraining configuration supports
both fine-tuned variant prediction and zero-shot functional inpainting. The
third expectation additionally requires sufficient flanking context,
train--test-matched corruption geometry, stable decoding, and a sensitive
functional evaluator. We test the masking comparisons directly and diagnose
the remaining conditions without retraining longer-context or kilobase-span
models.

\subsection{Density-Optimized Absorbing Diffusion}

We train ModernBERT with an absorbing discrete diffusion objective \citep{nie2025large}. 
For each training example, a masking ratio $r$ determines the target corruption level, and the model reconstructs the masked tokens from the remaining context. 
Given a masked set $M$, the training objective is
\begin{equation}
\mathcal{L}
=
\mathbb{E}_{r,M}
\left[
-\frac{1}{|M|}
\sum_{i\in M}
\log p_{\theta}(x_i\mid X_{\setminus M},r)
\right].
\end{equation}

\begin{figure}[t]
    \centering
    \includegraphics[width=\linewidth]{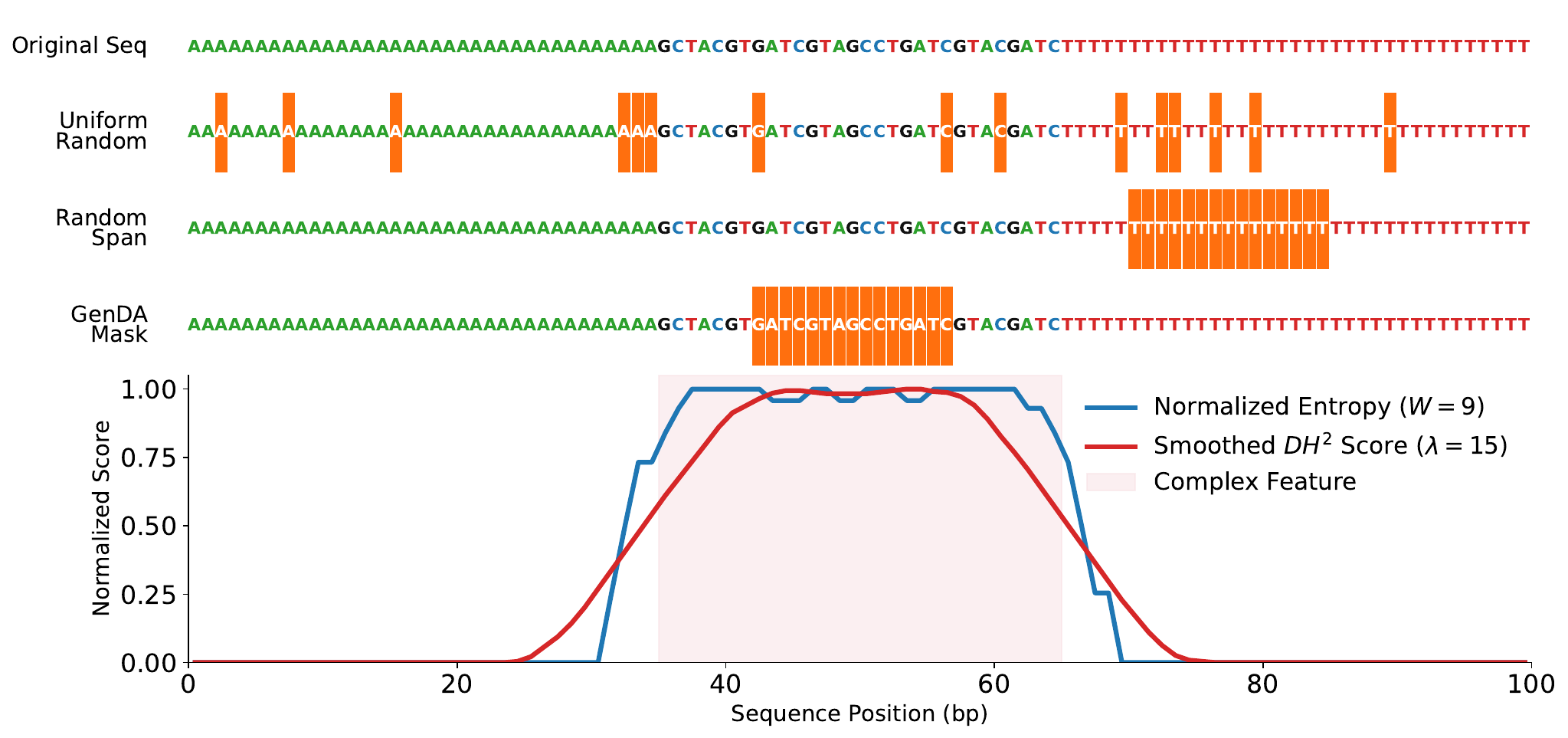}
    \caption{GenDA density-guided spans compared with independent uniform
    masking and Random Span.}
    \label{fig:adaptive_span}
\end{figure}

Rather than sampling independent masked positions, GenDA corrupts contiguous
spans whose locations are biased toward nucleotide-dense, locally complex
regions. Figure~\ref{fig:adaptive_span} illustrates the masking strategies;
complete pseudocode appears in \ref{alg:adaptive_span_masking}.

\begin{algorithm}[h]
\small
\caption{Adaptive density-optimized span masking}
\label{alg:adaptive_span_masking}
\begin{algorithmic}[1]
\Require Sequence $x$, window $W$, mean span length $\lambda$, mask ratio $r$
\State $M\gets\mathbf{0}$
\State Compute $H$, $D$, and $\Omega=\operatorname{AvgPool}_{K}(D\odot H^2)$
\State Zero invalid entries of $\Omega$ and normalize; $B\gets\lfloor rN_{\mathrm{valid}}\rfloor$
\While{$\sum_i M_i<B$}
    \State Sample $p\sim\operatorname{Categorical}(\Omega)$ and
    $\ell\sim\operatorname{Clip}(\operatorname{Poisson}(\lambda),1,300)$
    \If{$x_{p:p+\ell}$ contains no protected tokens}
        \State $M_{p:p+\ell}\gets1$
    \EndIf
\EndWhile
\State \Return $M$
\end{algorithmic}
\end{algorithm}

\paragraph{Progressive corruption curriculum.}
The masking ratio shifts from high to low across three stages: global
scaffolding uses $r\sim\mathcal{U}(0.60,1.00)$, structural reconstruction uses
$r\sim\mathcal{U}(0.20,0.70)$, and high-resolution refinement uses
$r\sim\mathcal{U}(0.05,0.30)$. In the final stage, $10\%$ of examples remain
fully masked ($r=1$) to preserve exposure to complete reconstruction.

\paragraph{Density-optimized span distribution.}
Genomic sequences exhibit substantial variation in local nucleotide
composition. Under position-uniform masking, compositionally simple and
complex regions are selected with equal probability. Because simple or
repetitive regions can often be reconstructed from short-range sequence
statistics, we bias span placement toward locally complex regions while using
nucleotide density to exclude assembly gaps and other non-ACGT positions.
Importantly, local entropy is used as a sequence-complexity heuristic and is
not assumed to identify biological function directly.

Let $\mathcal{A}=\{\mathrm{A},\mathrm{C},\mathrm{G},\mathrm{T}\}$ and let
$c_i^{(b)}$ count base $b$ in a width-$W$ window centered at $i$. With
$N_i=\sum_{b\in\mathcal{A}}c_i^{(b)}$ and
$p_i^{(b)}=c_i^{(b)}/(N_i+\epsilon)$, we define
\begin{equation}
H_i=-\sum_{b\in\mathcal{A}}p_i^{(b)}\log(p_i^{(b)}+\epsilon),
\qquad D_i=\frac{N_i}{W},
\qquad w_i=D_iH_i^2.
\end{equation}
We smooth at approximately the expected span scale and normalize valid starts:
\begin{equation}
K=2\left\lceil\frac{\lambda-1}{2}\right\rceil+1,
\quad \Omega=\operatorname{AvgPool}_K(w),
\quad P_{\mathrm{mask}}(i)=\frac{\Omega_i}{\sum_j\Omega_j}.
\end{equation}
Span starts follow $p\sim\operatorname{Categorical}(P_{\mathrm{mask}})$ and
lengths follow
$\ell\sim\operatorname{Clip}(\operatorname{Poisson}(\lambda),1,300)$.
Sampling continues until reaching $\lfloor rN_{\mathrm{valid}}\rfloor$ masked
nucleotides; the final span may slightly exceed this budget.

\paragraph{Rationale for structured corruption.}
Independent token masking can leave most of a local sequence pattern visible
while hiding only a few constituent nucleotides, allowing reconstruction from
immediately adjacent tokens. Contiguous span corruption removes neighboring
positions jointly and therefore requires the model to integrate information
over a broader context. Density guidance further increases the frequency
with which compositionally complex regions are reconstructed as coherent
units rather than as isolated nucleotides. Individual spans are capped at
300 bp; the objective therefore targets local-to-intermediate-scale
reconstruction and does not imply training on multi-kilobase contiguous gaps.
In the final configuration, 70\% of examples use density-guided contiguous
spans, while 30\% use independent position-uniform 1-mer masking. Thus, GenDA
retains limited exposure to token-level corruption, whereas the uniform
ablation applies it to every training example.

\subsection{Iterative Decoding}

For sequence reconstruction, GenDA begins from a task-defined set of masked
positions and iteratively refines its predictions. At step $t$, the model
predicts categorical nucleotide distributions for the masked positions.
Predictions are ranked by confidence, and the cumulative fraction retained is
determined by
\begin{equation}
\alpha_t
=
\sin\left(
\frac{t+1}{T}\frac{\pi}{2}
\right),
\qquad 0\leq t<T.
\end{equation}
Low-confidence positions are returned to \texttt{[MASK]} for subsequent
refinement, while high-confidence predictions are retained. The number of
refinement steps is adapted to the initial missing-region length:
\begin{equation}
T
=
\max\left(
\left\lfloor\frac{L_{\mathrm{gap}}}{30}\right\rfloor,
5
\right).
\end{equation}
This decoding procedure modifies only the inference trajectory and requires
no additional training. Unless stated otherwise, inpainting uses this
gap-adaptive schedule.

\section{Evaluation and Results}
\label{sec:experiments}

We first measure variant-effect prediction using ClinVar, then separately
evaluate functional generation using a matched-context AlphaGenome stress
test. GenDA and the AR Llama baseline contain 202M and 151M parameters,
respectively, and use identical supervised splits for ClinVar. Evo~2 is
evaluated zero-shot. Full data, training, and ablation details appear in
Appendix~\ref{app:experimental_setup}.

\subsection{Strong Fine-Tuned Variant Prediction}
\label{sec:clinvar}

\paragraph{Setup.}
We construct a high-confidence binary benchmark from ClinVar
\citep{clinvar_website}, retaining expert-reviewed SNVs labeled
\emph{Pathogenic} or \emph{Benign} and removing variants of uncertain
significance. Each variant is represented by a centered 256-bp sequence.
GenDA and AR Llama undergo full-parameter supervised fine-tuning on identical
gene-aware splits, whereas Evo~2 is evaluated zero-shot.

GenDA scores each variant using the inverted log-likelihood ratio
\begin{equation}
s(x)
=
\log p_\theta(x_{\mathrm{ref}}\mid x_{\setminus i})
-
\log p_\theta(x_{\mathrm{alt}}\mid x_{\setminus i}),
\end{equation}
where larger values indicate that the alternative allele is less compatible
with its surrounding context. We report AUROC by functional region and over
the pooled benchmark.

\paragraph{Results.}
Table~\ref{tab:snv} presents the main discriminative result. After supervised
fine-tuning, GenDA reaches an overall AUROC of \(0.774\), exceeding the
fine-tuned AR Llama baseline by 0.103 and zero-shot Evo~2 by 0.041. This
establishes that the complete GenDA pretraining and fine-tuning configuration
supports strong downstream variant-effect classification. It does not by
itself identify whether the gain arises from bidirectional attention, the
diffusion objective, the curriculum, the backbone, or entropy-guided masking.
The following ablations narrow this attribution.

\begin{table*}[t]
\centering
\caption{ClinVar SNV AUROC. GenDA w/o Curriculum uses the same masking policy
without the progressive corruption curriculum. Evo~2 is evaluated zero-shot;
AR Llama and both GenDA configurations undergo supervised fine-tuning. Bold
denotes the best result in each row.}
\label{tab:snv}
\resizebox{0.68\textwidth}{!}{%
\begin{tabular}{lcccc}
\toprule
\textbf{Region}
& \textbf{Evo~2}
& \textbf{AR Llama}
& \textbf{GenDA w/o Curriculum}
& \textbf{GenDA} \\
\midrule
Enhancer
& \textbf{0.747} & 0.662 & 0.704 & 0.746 \\
Exon
& 0.699 & 0.665 & 0.722 & \textbf{0.757} \\
Intron
& 0.719 & 0.663 & 0.739 & \textbf{0.773} \\
Promoter
& 0.768 & 0.694 & 0.730 & \textbf{0.778} \\
\midrule
Overall
& 0.733 & 0.671 & 0.743 & \textbf{0.774} \\
\bottomrule
\end{tabular}%
}
\end{table*}

\subsection{Entropy Guidance Does Not Beat Random Spans}
\label{sec:main_ablations}

Exclusive independent 1-mer masking collapses during unconditional generation
in the evaluated run (Appendix~\ref{app:uniform}), whereas GenDA's span-dominated mixture does not.

Density guidance does not yet provide a uniform advantage over random spans.
At the matched 12-layer scale, a random-span variant achieves an overall
ClinVar AUROC of \(0.777\), compared with \(0.774\) for GenDA. The inpainting
comparison is similarly task-dependent: GenDA obtains higher median restoration
gain for promoters, exon boundaries, and intron boundaries, whereas random
spans perform better on enhancers.

A plausible interpretation is that ClinVar SNV scoring is comparatively local:
each allele is evaluated within a centered 256-bp window, and many pathogenic
effects are detectable from nearby coding or regulatory context. Random
contiguous spans already train reconstruction at this local scale, so
entropy-guided placement may add little benefit for this benchmark. This
interpretation is task-specific and does not imply that all variant effects are
local.

Thus, the span-dominated mixture avoids the observed independent-masking
collapse, but density-guided placement is not uniformly superior to Random
Span. Complete masking, scaling, and inpainting results appear in
Appendix~\ref{app:masking_scaling}.

\subsection{Discriminative Success Does Not Transfer}
\label{sec:functional_inpainting}

\paragraph{Evaluation objective.}
We test whether GenDA's fine-tuned discriminative performance transfers to
zero-shot functional sequence reconstruction. The loci remain within the human genomic domain, but
corruption geometry shifts: training distributes the masking budget across
spans capped at 300 bp, whereas evaluation removes one contiguous region of up
to 3,500 bp. The experiment therefore measures both in-distribution short-gap
reconstruction and increasingly strong length extrapolation.

\paragraph{Setup.}
We sample 200 non-overlapping held-out sites comprising 50 promoters,
50 enhancers, 50 exons, and 50 introns. For each site, we reconstruct gaps
from 50 to 3,500 bp using three samples per gap length. GenDA conditions on
both flanks, whereas AR Llama and Evo~2 generate from the left context. D3LM
is included as an external diffusion reference, although its shorter and more
curated training distribution prevents a controlled comparison.

Each reconstructed gap is inserted into the same 524,288-bp hg38 context and
evaluated against the native AlphaGenome prediction
\citep{avsec2026advancing}. We measure promoter CAGE, enhancer H3K27ac, and
splice-site usage at exon and intron boundaries.

To account for locus- and gap-dependent AlphaGenome sensitivity, we compare
each reconstruction with controls obtained by shuffling the native gap while
exactly preserving its 3-mer composition. Restoration gain is
\begin{equation}
\operatorname{RG}
=
100\left(
1-\frac{D_{\mathrm{model}}}{D_{\mathrm{control}}}
\right),
\end{equation}
where zero denotes parity with the corrupted control and positive values
indicate improved restoration. RG is unbounded below; for example, \(-100\)
indicates twice the restoration distance of the control. We report site-level
medians and 95\% confidence intervals from 10,000 bootstrap replicates.
Complete metric and eligibility definitions are provided in
Appendix~\ref{app:alphagenome_evaluation}.

\begin{table*}[t]
\centering
\caption{\textbf{AlphaGenome functional inpainting across gap lengths.}
Entries are median site-level restoration gain averaged across evaluated gap
lengths, with 95\% bootstrap confidence intervals shown as subscripts. Higher
is better; zero denotes parity with the 3-mer-preserving corrupted control.}
\label{tab:alphagenome_rg}
\resizebox{0.8\textwidth}{!}{%
\begin{tabular}{lcccc}
\toprule
\textbf{Model}
& \textbf{Promoter CAGE}
& \textbf{Enhancer H3K27ac}
& \textbf{Exon usage}
& \textbf{Intron usage} \\
\midrule
Evo~2
& $-13.6_{\scriptscriptstyle[-16.8,-9.0]}$
& $-23.0_{\scriptscriptstyle[-37.5,-12.6]}$
& $-2.8_{\scriptscriptstyle[-12.1,-0.1]}$
& $0.0_{\scriptscriptstyle[-1.6,0.0]}$ \\
AR Llama
& $-14.8_{\scriptscriptstyle[-18.7,-12.1]}$
& $-31.1_{\scriptscriptstyle[-37.9,-25.2]}$
& $-6.2_{\scriptscriptstyle[-15.7,-0.2]}$
& $0.0_{\scriptscriptstyle[-4.7,0.0]}$ \\
D3LM
& $-41.5_{\scriptscriptstyle[-73.9,-20.3]}$
& $-84.1_{\scriptscriptstyle[-118.1,-60.8]}$
& $-12.0_{\scriptscriptstyle[-28.8,0.1]}$
& $-2.2_{\scriptscriptstyle[-8.3,0.1]}$ \\
\textbf{GenDA}
& $-37.9_{\scriptscriptstyle[-65.9,-27.2]}$
& $-130.8_{\scriptscriptstyle[-206.6,-62.5]}$
& $-17.0_{\scriptscriptstyle[-22.9,0.1]}$
& $0.1_{\scriptscriptstyle[-0.9,0.2]}$ \\
\midrule
Eligible sites & 43/50 & 25/50 & 47/50 & 45/50 \\
\bottomrule
\end{tabular}%
}
\end{table*}

\paragraph{Observed outcome.}
Contrary to our primary generative hypothesis, no evaluated model consistently
improves over the composition-preserving control. Promoter and enhancer RG is
negative for every model, and GenDA is particularly weak on enhancers despite
its strong ClinVar performance. Random Span changes the ordering across
modalities, but its confidence intervals overlap GenDA's in all four categories
(Appendix~\ref{app:random_span_inpainting}).

Stratification by gap length shows that the failure is not confined to
long-range extrapolation (Appendix~Table~\ref{tab:alphagenome_gap_ranges}).
Promoter and enhancer RG remain negative for every model even at 50--500 bp.
Length effects are nevertheless modality-specific: GenDA's enhancer RG
deteriorates sharply for 2,000--3,500-bp gaps, whereas splice-usage RG remains
near zero across ranges. Because RG can be unstable when the control distance
is small, the extreme enhancer ratio should not be read as a literal effect
size. Long gaps and limited flanking context may aggravate failure, but they do
not explain its presence at shorter scales.

\section{Diagnosing the Failed Premise}

\subsection{Limited Context and Corruption-Geometry Shift}

Two design choices jointly constrain the inpainting experiment. First, 1-mer
tokenization preserves single-nucleotide resolution but limits a 4,096-token
window to 4,096 bp. At a 3,500-bp gap, fewer than 600 observed flank
nucleotides remain. Second, each training span has mean length 100 bp and is
clipped at 300 bp. Multiple such spans can produce a high total masking ratio
while leaving internal observed anchors, which differs from one uninterrupted
3,500-bp gap. For descriptive analysis, we group gaps into short
(50--500 bp), medium (800--1,500 bp), and long (2,000--3,500 bp) ranges
(Appendix~Table~\ref{tab:alphagenome_gap_ranges}); only the 50- and 200-bp
gaps fall within the 300-bp training-span cap.

These are plausible boundary conditions rather than verified causes: we did
not retrain with a longer physical context, coarser tokenization, or
kilobase-scale single spans. Coarser tokens could cover more bases at a fixed
token budget, but would change the learning problem and complicate the
single-nucleotide likelihood used for ClinVar. The current results therefore
show that bidirectional access to the \emph{available} flanks is insufficient,
not that longer-context bidirectional modeling cannot help. AlphaGenome itself
is non-causal: after generation, every completed 524,288-bp sequence is
evaluated identically; causality differs only during gap generation.

\subsection{Sequence Complexity Is Not Functional Importance}

The weakest assumption in the GenDA premise was that reconstruction difficulty
would indicate where training capacity should be allocated. GenDA uses
$D_iH_i^2$ to favor nucleotide-dense, compositionally diverse regions, but this
hardness heuristic does not identify transcription start sites, enhancer
activity, splice regulation, or positions to which AlphaGenome is sensitive.
The random-span ablation makes this
limitation empirical: random spans slightly exceed GenDA on pooled ClinVar
(0.777 versus 0.774) and improve enhancer RG substantially
($-58.4$ versus $-130.8$), although confidence intervals overlap. Density
guidance performs better on the other three restoration categories. The prior
therefore changes which failures occur rather than delivering a uniform
functional advantage. ClinVar may also be unusually favorable to random spans:
the benchmark scores one SNV in a 256-bp window, so nearby sequence context can
be sufficient for many examples. Contiguous masking itself may provide the
relevant local reconstruction pressure, leaving limited room for the
span-start prior to improve this task.

\subsection{Absolute Fidelity Can Conceal Control-Level Performance}

Enhancer H3K27ac fidelity appears high for every model (90.1--95.7; Appendix
Table~\ref{tab:alphagenome_fidelity}), yet all enhancer RG values are negative.
Most of the 524-kb input context is unchanged, H3K27ac profiles are spatially
smooth, and some channels or loci may respond weakly to the replaced gap. A
model can therefore remain close to the native prediction without improving on
a simple composition-preserving corruption. The shuffled control is what turns
apparently strong absolute fidelity into the more informative conclusion that
the learned generators add no consistent restoration value.

RG has its own limitation. Because it divides by the control distance, it can
become extremely negative when the control changes AlphaGenome only slightly.
We exclude control distances below 0.01 and report medians and bootstrap
intervals, but values just above the threshold can still be unstable. We
therefore report absolute fidelity alongside RG and recommend also inspecting
the non-ratio difference $D_{\mathrm{control}}-D_{\mathrm{model}}$ in future
evaluations. No single predictor-based score should be treated as experimental
validation.

\subsection{Inference-Time Rules Redistribute Rather Than Repair Failure}

BioCheck re-masks low-confidence positions flagged by simplified promoter,
reading-frame, splice, motif, and GC-content rules. It improves enhancer RG
from $-130.8$ to $-42.6$ but degrades promoter RG from $-37.9$ to $-69.4$ and
leaves splice usage essentially unchanged (Appendix~\ref{app:biocheck}). The
heuristic therefore shifts performance between modalities instead of producing
general functional restoration. This is a caution against evaluating a
biological constraint on only the assay it directly favors.

\section{Actionable Takeaways and Limitations}

Although our experiments center on GenDA, the observed failure modes suggest a
broader evaluation protocol for genomic sequence generators. The goal is not
to discourage functional generation, but to identify inexpensive checks that
can falsify weak claims before researchers invest in large-scale training,
predictor inference, or experimental validation.

\begin{enumerate}
    \item \textbf{Match evidence to the claimed capability.}
    Variant prediction, sequence likelihood, native-profile similarity, and
    functional generation measure different capabilities. Success on one can
    motivate, but should not substitute for, direct evaluation of another.

    \item \textbf{Report physical context and corruption geometry together.}
    A nominally long token window may cover little DNA under 1-mer tokenization,
    and matching total masking ratio does not match a single long missing
    interval. Gap length, contiguity, physical coverage, and remaining flank
    context should accompany inpainting results.

    \item \textbf{Test evaluator sensitivity before evaluating generators.}
    For each locus, gap length, and assay, first measure how much a
    composition-preserving perturbation changes the predictor. If the control
    barely affects the output, high fidelity is not evidence of successful
    reconstruction and ratio-based improvement becomes unstable. This check can
    prevent expensive evaluation on insensitive locus--assay combinations. Our
    exact 3-mer-preserving shuffles provide this calibration while retaining
    local sequence composition.

    \item \textbf{Stratify by sequence scale and functional readout.}
    Aggregate scores concealed distinct failure modes in our evaluation:
    promoter and enhancer restoration was already poor at short gaps, long gaps
    selectively amplified GenDA's enhancer failure, and splice readouts remained
    largely insensitive. Reporting gap regimes and modalities separately helps
    distinguish general generation failure, scale-specific degradation, and
    evaluator insensitivity.

    \item \textbf{Treat biological guidance as multi-objective.}
    A rule or reward that improves one predictor output may degrade another.
    Guidance mechanisms should be evaluated across all relevant modalities and
    reported as trade-offs rather than as general biological improvement.
\end{enumerate}

Together, these checks can reject weak generative claims before investment in
larger models, additional predictors, or experimental validation.

Our study cannot separate the effects of the 4,096-bp context, 1-mer
tokenization, and 300-bp span cap because we did not retrain matched longer-
context, coarser-token, or kilobase-span models. AlphaGenome does not measure
experimental activity, general sequence realism, or diversity. D3LM and Evo~2
are descriptive external references with different corpora and capacities,
and the 200-site panel gives wide intervals for splice usage. The empirical
claims therefore remain limited to the evaluated models, predictor, and loci;
the proposed protocol is diagnostic guidance rather than a universal
guarantee. Future work may isolate context coverage, tokenization, and span
geometry before expanding to additional predictors or experimental validation.

\section{Conclusion}

GenDA was developed under the hypothesis that bidirectional denoising,
contiguous corruption, and entropy-guided span placement would jointly benefit
both genomic prediction and generation. The results provide only partial
support. The complete system performs strongly on ClinVar after supervised
fine-tuning, and exclusive independent 1-mer masking collapses in the evaluated
generation run. However, Random Span slightly exceeds GenDA on pooled ClinVar,
so the entropy prior is not validated, and neither masking strategy produces
reliable zero-shot functional inpainting.

The most informative result is therefore the failed transfer of the original
premise: compositionally complex sequence is not necessarily functionally
important, and strong fine-tuned variant prediction does not establish that a
pretrained model has learned a useful conditional sequence distribution.
Limited physical context, corruption-geometry mismatch, and evaluator
sensitivity may contribute to the observed failure, but none is established as
its sole cause. Future genomic generators should test masking priors against
simple random-span controls and evaluate discriminative and generative
capabilities separately before attributing success to a biologically motivated
pretraining design.

\section*{Acknowledgement}
This work is partly supported by the Federal Ministry of Research, Technology and Space in DAAD project 57616814 (SECAI, School of Embedded Composite AI, https://secai.org/) as part of the program Konrad Zuse Schools of Excellence in Artificial Intelligence.

\bibliographystyle{plainnat} 
\setcitestyle{authoryear,open={(},close={)}}
\bibliography{main}


\appendix

\section{Ablation Studies}

\subsection{Exclusive Uniform 1-mer Masking Collapses}
\label{app:uniform}

To test whether independent token corruption can serve as the sole training
policy, we replace GenDA's masking mixture with independent position-uniform
1-mer masking for every training example. The ModernBERT backbone, training
data, corruption-ratio curriculum, optimization hyperparameters, and
computational budget remain unchanged.

\begin{figure}[t]
    \centering
    \includegraphics[width=\linewidth]
        {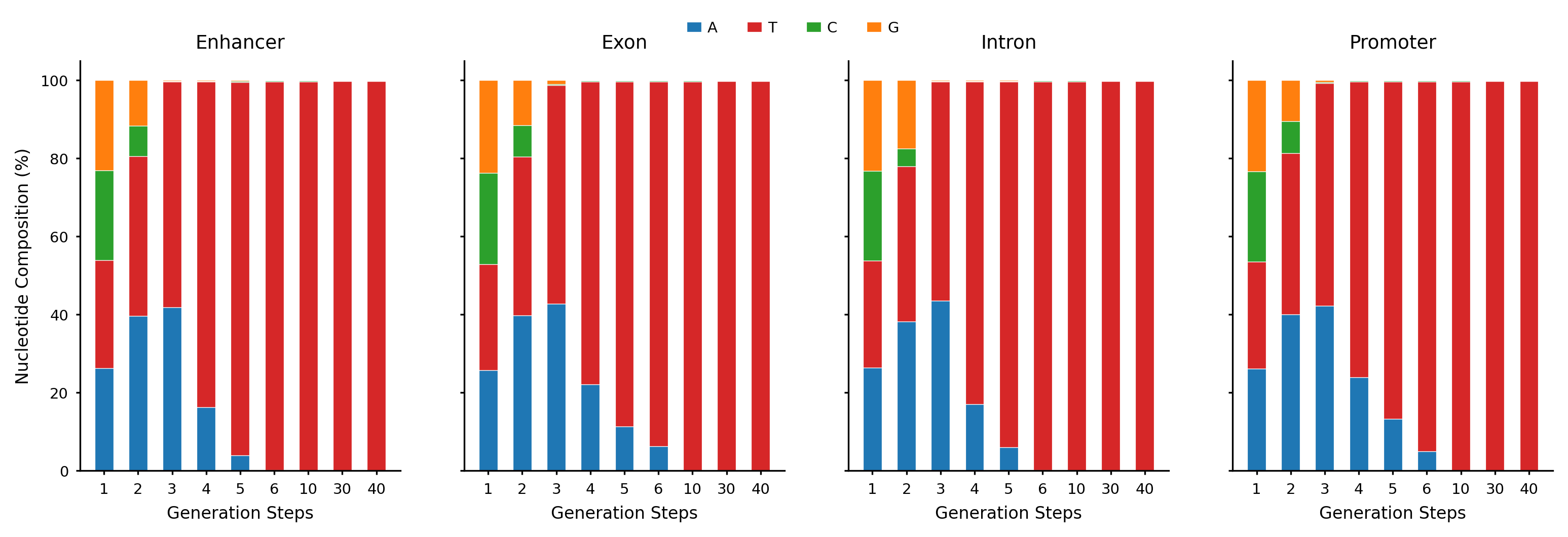}
    \caption{\textbf{Collapse under exclusive independent uniform 1-mer
    masking.} Nucleotide frequencies across unconditional denoising steps.
    The model rapidly concentrates on a single nucleotide despite using the
    same backbone, data, curriculum, and computational budget as GenDA.}
    \label{fig:ablation_random}
\end{figure}

During unconditional generation from a fully masked sequence, the evaluated
uniform-masking model rapidly concentrates its predictions on a single
nucleotide and remains collapsed throughout the remaining denoising steps
(Fig.~\ref{fig:ablation_random}).

This result does not imply that a limited uniform component is harmful:
GenDA uses independent 1-mer masking for 30\% of training examples. It also
does not imply that uniform masked diffusion must collapse under other
datasets, tokenizations, model initializations, or training runs. Instead, the
experiment shows that exclusive independent 1-mer corruption is insufficient
in the evaluated GenDA setting.

\subsection{Random-Span and Scaling Ablations}
\label{app:masking_scaling}

Table~\ref{tab:ablation_masking_vs_size} reports region-specific ClinVar AUROC
after each successive curriculum stage. These are sequential checkpoints
rather than compute-matched standalone models.

At the final stage, random spans slightly outperform the 12-layer GenDA model
across all four regions, reaching an overall AUROC of \(0.777\) compared with
\(0.774\). Thus, the current experiment does not establish density-guided
placement as superior to random span placement.

Increasing model depth produces a more consistent improvement. The 24-layer
GenDA-large model achieves the strongest final-stage result in all four
regions and reaches an overall AUROC of \(0.786\). Across all intermediate
checkpoints, Random Span P2 remains marginally best on exons
(\(0.764\) versus \(0.762\)). Because masking policy and capacity are not
jointly controlled, the scaling result shows that the GenDA masking mixture is
compatible with additional capacity but does not establish a causal advantage
for density guidance.

\begin{table*}[t]
\centering
\caption{Region-specific ClinVar SNV AUROC across masking strategies, model
sizes, and successive curriculum stages. P1--P3 are sequential checkpoints
rather than compute-matched standalone models. GenDA-large uses twice the
depth of GenDA. Higher is better.}
\label{tab:ablation_masking_vs_size}
\resizebox{0.8\textwidth}{!}{%
\begin{tabular}{lccc ccc ccc}
\toprule
\textbf{Region}
& \multicolumn{3}{c}{\textbf{Random Span}}
& \multicolumn{3}{c}{\textbf{GenDA}}
& \multicolumn{3}{c}{\textbf{GenDA-large}} \\
\cmidrule(lr){2-4}
\cmidrule(lr){5-7}
\cmidrule(lr){8-10}
& P1 & P2 & P3
& P1 & P2 & P3
& P1 & P2 & P3 \\
\midrule
Enhancer
& 0.737 & 0.741 & 0.747
& 0.697 & 0.716 & 0.746
& 0.740 & 0.757 & \textbf{0.768} \\
Exon
& 0.758 & \textbf{0.764} & 0.758
& 0.727 & 0.740 & 0.757
& 0.746 & 0.757 & 0.762 \\
Intron
& 0.759 & 0.777 & 0.778
& 0.723 & 0.739 & 0.773
& 0.765 & 0.777 & \textbf{0.787} \\
Promoter
& 0.765 & 0.775 & 0.785
& 0.720 & 0.757 & 0.778
& 0.760 & 0.784 & \textbf{0.787} \\
\midrule
Overall
& 0.770 & 0.779 & 0.777
& 0.735 & 0.750 & 0.774
& 0.765 & 0.776 & \textbf{0.786} \\
\bottomrule
\end{tabular}%
}
\end{table*}

\subsection{Random-Span Functional Inpainting}
\label{app:random_span_inpainting}

We additionally compare GenDA with random-span masking under the AlphaGenome
inpainting evaluation. Random Span uses the same Poisson span-length
distribution as GenDA but samples span starts uniformly. 

GenDA obtains higher median restoration gain for promoters, exon boundaries,
and intron boundaries, while Random Span performs substantially better on
enhancers. The confidence intervals overlap in all four categories, so the
results should be interpreted as task-dependent trends rather than
statistically established differences.

\begin{table*}[t]
\centering
\caption{AlphaGenome inpainting ablation comparing Random Span with GenDA.
Entries are median restoration gain, with 95\% bootstrap confidence intervals
shown as subscripts. Higher is better.}
\label{tab:random_span_inpainting}
\resizebox{0.8\textwidth}{!}{%
\begin{tabular}{lcccc}
\toprule
\textbf{Model}
& \textbf{Promoter CAGE}
& \textbf{Enhancer H3K27ac}
& \textbf{Exon usage}
& \textbf{Intron usage} \\
\midrule
Random Span
& $-56.8_{\scriptscriptstyle[-130.7,-34.1]}$
& $\mathbf{-58.4}_{\scriptscriptstyle[-89.4,-33.8]}$
& $-19.3_{\scriptscriptstyle[-42.9,0.3]}$
& $-0.2_{\scriptscriptstyle[-10.1,0.3]}$ \\
\textbf{GenDA}
& $\mathbf{-37.9}_{\scriptscriptstyle[-65.9,-27.2]}$
& $-130.8_{\scriptscriptstyle[-206.6,-62.5]}$
& $\mathbf{-17.0}_{\scriptscriptstyle[-22.9,0.1]}$
& $\mathbf{0.1}_{\scriptscriptstyle[-0.9,0.2]}$ \\
\bottomrule
\end{tabular}%
}
\end{table*}

\subsection{Exploratory BioCheck Decoding}
\label{app:biocheck}

BioCheck is an inference-only heuristic that adds rule-flagged positions to
the confidence-based re-masking set. It does not directly edit nucleotides or
guarantee that a flagged pattern is corrected. The rules examine simplified
promoter motifs, reading frames, splice-like patterns, transcription-factor
motifs, and category-specific GC-content ranges.

\begin{table*}[t]
\centering
\caption{Exploratory BioCheck ablation. Entries are median AlphaGenome
restoration gain, with 95\% bootstrap confidence intervals shown as
subscripts. Higher is better.}
\label{tab:biocheck_ablation}
\resizebox{0.8\textwidth}{!}{%
\begin{tabular}{lcccc}
\toprule
\textbf{Model}
& \textbf{Promoter CAGE}
& \textbf{Enhancer H3K27ac}
& \textbf{Exon usage}
& \textbf{Intron usage} \\
\midrule
GenDA
& $-37.9_{\scriptscriptstyle[-65.9,-27.2]}$
& $-130.8_{\scriptscriptstyle[-206.6,-62.5]}$
& $-17.0_{\scriptscriptstyle[-22.9,0.1]}$
& $0.1_{\scriptscriptstyle[-0.9,0.2]}$ \\
GenDA+BioCheck
& $-69.4_{\scriptscriptstyle[-122.0,-40.2]}$
& $-42.6_{\scriptscriptstyle[-53.7,-36.1]}$
& $-17.6_{\scriptscriptstyle[-27.7,0.4]}$
& $0.1_{\scriptscriptstyle[-0.2,0.3]}$ \\
\bottomrule
\end{tabular}%
}
\end{table*}

BioCheck improves enhancer restoration but degrades promoter restoration and
has little effect on splice usage. We therefore interpret it as a
modality-specific heuristic intervention rather than evidence of improved
biological validity.

\section{AlphaGenome Evaluation Details}
\label{app:alphagenome_evaluation}

\paragraph{Protocol.}
We evaluate 200 held-out sites (50 promoters, 50 enhancers, 50 exons, and
50 introns) using gaps of
$g\in\{50,200,500,800,1000,1500,2000,2500,3000,3500\}$ bp and three
reconstructions per site--model--gap combination. Each reconstruction is
inserted into the same 524,288-bp hg38 context as the native sequence, such
that only the generated gap differs. We compare log-transformed promoter CAGE
and enhancer H3K27ac profiles over their annotated intervals and raw
splice-site usage at exon and intron boundaries. All non-padding AlphaGenome
channels are macro-averaged; the evaluation is therefore cross-condition rather
than tissue-specific.

\paragraph{Metrics and controls.}
Let $\Omega_c$ and $\mathcal{T}_c$ denote the evaluated positions and channels
for category $c$. We use $z=\log(1+y)$ for CAGE and H3K27ac and $z=y$ for
splice usage. For reconstruction $i$,
\begin{equation}
D_c^{(i)}
=
\frac{1}{|\mathcal{T}_c|}
\sum_{t\in\mathcal{T}_c}
\frac{\sum_{u\in\Omega_c}
|z_{u,t}^{\mathrm{pred},(i)}-z_{u,t}^{\mathrm{native}}|}
{\sum_{u\in\Omega_c}
(z_{u,t}^{\mathrm{pred},(i)}+z_{u,t}^{\mathrm{native}})+10^{-6}}.
\label{eq:appendix_restoration_distance}
\end{equation}
We convert this normalized distance into absolute restoration fidelity,
\begin{equation}
F_c^{(i)}=100\left(1-D_c^{(i)}\right),
\label{eq:appendix_restoration_fidelity}
\end{equation}
where $F_c^{(i)}=100$ denotes an AlphaGenome prediction identical to the native
profile and $F_c^{(i)}=0$ denotes maximal normalized disagreement. Unlike RG,
absolute fidelity does not depend on a corrupted control.

For every site and gap, we construct three controls by shuffling the native gap
while exactly preserving its 3-mer composition. With
$\overline D_{c,\mathrm{ctrl}}$ denoting their mean distance, restoration gain
is
\begin{equation}
\operatorname{RG}_c^{(i)}
=100\left(1-
\frac{D_c^{(i)}}{\overline D_{c,\mathrm{ctrl}}}\right).
\end{equation}
Positive RG indicates improvement over the corrupted control. We exclude cases
with $\overline D_{c,\mathrm{ctrl}}<0.01$ because the ratio is unstable when
the control has little predicted effect. For both fidelity and RG, values are
averaged across the three reconstructions and then across gaps within each
site; RG summaries additionally apply the control-distance eligibility rule.
We report site-level medians and percentile 95\% confidence intervals from
10,000 bootstrap resamples of sites.

\paragraph{Interpretation.}
Because AlphaGenome similarity does not directly measure sequence realism,
diversity, or low-complexity collapse, the reported scores are interpreted only
as predictor-based restoration. The ratio form of RG is also sensitive to small
control distances despite the eligibility threshold. These results therefore
do not constitute experimental evidence of biological activity or general
sequence quality.

\paragraph{Gap-range aggregation.}
To test whether the aggregate result is driven by long out-of-distribution
gaps, we additionally group gaps into short (50--500 bp), medium
(800--1,500 bp), and long (2,000--3,500 bp) ranges. Within each site and range,
we first average the three reconstructions at each gap length and then average
across gap lengths; confidence intervals bootstrap sites. Table~\ref{tab:alphagenome_gap_ranges}
shows that negative promoter and enhancer RG is already present in the short
range. The large negative long-gap enhancer ratios, particularly for GenDA,
should be interpreted together with the ratio-instability caveat above.

\begin{table*}[t]
\centering
\caption{AlphaGenome restoration gain stratified by gap range. Entries report
median site-level RG with 95\% site-bootstrap confidence intervals. Short:
50--500 bp; medium: 800--1,500 bp; long: 2,000--3,500 bp. Random Span is the
GenDA masking ablation from Appendix~\ref{app:random_span_inpainting}.}
\label{tab:alphagenome_gap_ranges}
{
\setlength{\tabcolsep}{3.5pt}
\renewcommand{\arraystretch}{1.08}
\resizebox{\textwidth}{!}{%
\begin{tabular}{llcccc}
\toprule
\textbf{Model} & \textbf{Gap range}
& \textbf{Promoter CAGE RG $\uparrow$}
& \textbf{Enhancer H3K27ac RG $\uparrow$}
& \textbf{Exon usage RG $\uparrow$}
& \textbf{Intron usage RG $\uparrow$} \\
\midrule
\multirow{3}{*}{Evo~2}
& Short  & $-23.6_{\;[-33.1,-14.9]}$ & $-30.0_{\;[-39.1,-13.3]}$ & $-3.5_{\;[-11.9,-0.2]}$ & $-0.0_{\;[-0.4,0.0]}$ \\
& Medium & $-3.4_{\;[-10.2,0.2]}$ & $-25.9_{\;[-33.5,-18.2]}$ & $-0.1_{\;[-5.1,0.0]}$ & $-0.0_{\;[-1.8,0.0]}$ \\
& Long   & $-8.4_{\;[-14.6,-3.8]}$ & $-15.3_{\;[-23.6,-8.6]}$ & $0.0_{\;[-4.0,0.0]}$ & $0.0_{\;[0.0,0.0]}$ \\
\midrule
\multirow{3}{*}{AR Llama}
& Short  & $-29.4_{\;[-33.5,-22.9]}$ & $-38.0_{\;[-53.4,-31.4]}$ & $-8.8_{\;[-25.0,-0.2]}$ & $-0.0_{\;[-3.6,0.0]}$ \\
& Medium & $-7.0_{\;[-12.5,-3.9]}$ & $-43.3_{\;[-50.8,-32.2]}$ & $-0.3_{\;[-7.1,-0.0]}$ & $0.0_{\;[-0.1,0.0]}$ \\
& Long   & $-7.3_{\;[-14.7,-4.9]}$ & $-28.0_{\;[-46.2,-21.4]}$ & $-0.0_{\;[-2.2,0.0]}$ & $0.0_{\;[0.0,0.1]}$ \\
\midrule
\multirow{3}{*}{D3LM}
& Short  & $-49.1_{\;[-67.9,-36.6]}$ & $-87.5_{\;[-104.5,-59.1]}$ & $-5.1_{\;[-31.4,-0.5]}$ & $-0.0_{\;[-6.8,0.0]}$ \\
& Medium & $-55.7_{\;[-95.2,-25.6]}$ & $-108.6_{\;[-139.4,-85.9]}$ & $0.0_{\;[-22.6,0.2]}$ & $0.0_{\;[-9.0,0.2]}$ \\
& Long   & $-36.3_{\;[-50.5,-19.3]}$ & $-97.1_{\;[-115.3,-77.1]}$ & $0.1_{\;[-18.8,0.3]}$ & $0.1_{\;[-0.2,0.2]}$ \\
\midrule
\multirow{3}{*}{\textbf{GenDA}}
& Short  & $-40.0_{\;[-62.0,-30.8]}$ & $-49.4_{\;[-66.3,-33.3]}$ & $-2.9_{\;[-16.7,-0.0]}$ & $0.0_{\;[-5.2,0.0]}$ \\
& Medium & $-42.7_{\;[-111.1,-22.8]}$ & $-102.6_{\;[-167.5,-60.4]}$ & $-0.2_{\;[-22.8,0.1]}$ & $0.1_{\;[-0.0,0.5]}$ \\
& Long   & $-37.2_{\;[-45.8,-21.4]}$ & $-277.9_{\;[-332.9,-221.2]}$ & $-1.2_{\;[-26.3,0.1]}$ & $0.1_{\;[-0.0,0.2]}$ \\
\midrule
\multirow{3}{*}{Random Span}
& Short  & $-35.7_{\;[-69.3,-24.9]}$ & $-48.4_{\;[-58.0,-30.7]}$ & $-7.0_{\;[-20.4,0.0]}$ & $-0.0_{\;[-4.3,0.0]}$ \\
& Medium & $-87.6_{\;[-168.5,-38.6]}$ & $-69.0_{\;[-84.6,-50.5]}$ & $0.0_{\;[-21.8,0.3]}$ & $0.0_{\;[-10.8,0.1]}$ \\
& Long   & $-80.6_{\;[-127.4,-52.3]}$ & $-71.6_{\;[-108.2,-49.3]}$ & $0.3_{\;[-44.6,1.1]}$ & $0.2_{\;[0.1,0.8]}$ \\
\bottomrule
\end{tabular}%
}
}
\end{table*}

\begin{table*}[t]
\centering
\caption{Absolute AlphaGenome restoration fidelity across all gap lengths.
Entries report median gap-averaged fidelity with 95\% site-bootstrap confidence
intervals. Higher values indicate closer agreement with the native AlphaGenome
prediction. High enhancer fidelity contrasts with negative control-normalized
restoration gain in Table~\ref{tab:alphagenome_rg}.}
\label{tab:alphagenome_fidelity}
\resizebox{0.9\textwidth}{!}{%
\begin{tabular}{lcccc}
\toprule
\textbf{Model}
& \textbf{Promoter CAGE}
& \textbf{Enhancer H3K27ac}
& \textbf{Exon usage}
& \textbf{Intron usage} \\
\midrule
Evo~2
& $67.9_{\;[56.8,73.5]}$
& $95.7_{\;[95.3,96.1]}$
& $35.5_{\;[20.1,67.7]}$
& $46.8_{\;[7.5,56.2]}$ \\
AR Llama
& $69.1_{\;[57.0,72.8]}$
& $95.2_{\;[94.8,95.5]}$
& $34.6_{\;[20.0,67.6]}$
& $47.5_{\;[7.6,57.2]}$ \\
D3LM
& $55.7_{\;[51.9,58.9]}$
& $92.8_{\;[92.4,93.2]}$
& $36.8_{\;[24.3,63.3]}$
& $44.8_{\;[8.8,56.6]}$ \\
GenDA
& $49.6_{\;[44.7,54.8]}$
& $90.1_{\;[88.6,91.9]}$
& $38.6_{\;[23.6,62.8]}$
& $47.7_{\;[10.0,56.4]}$ \\
GenDA+BioCheck
& $42.6_{\;[39.2,44.4]}$
& $95.1_{\;[94.6,95.7]}$
& $39.7_{\;[26.8,58.6]}$
& $45.9_{\;[12.9,57.5]}$ \\
\bottomrule
\end{tabular}%
}
\end{table*}

\section{Detailed Experimental Setup}
\label{app:experimental_setup}

\subsection{Data Curation and Splitting}
We extract diverse functional anchors—promoters (TSS $\pm$ 2000 bp), exons, enhancers, introns, and repeats—from the human reference genome (hg38) using GENCODE GTF and standard peak annotations. Sequences containing $>5\%$ uncalled bases (\texttt{N}) are discarded to maintain data quality. To prevent homologous data leakage during evaluation, we utilize a strict gene-aware split partitioned by \texttt{gene\_id}, yielding approximately 80\% training, 10\% validation, and 10\% test sets. The exact global distribution of these anchors is detailed in Table~\ref{tab:dataset_splits}.

\begin{table*}[t]
\caption{\textbf{Global Dataset Split Summary.} The dataset comprises $\sim$3.2 million genomic anchors partitioned via a strict gene-aware split to prevent structural leakage.}
\label{tab:dataset_splits}
\centering
\resizebox{0.7\textwidth}{!}{%
\begin{tabular}{@{}l r r r r r@{}}
\toprule
\textbf{Category} & \textbf{Train} & \textbf{Validation} & \textbf{Test} & \textbf{Total} & \textbf{\% of Total} \\
\midrule
Exon & 869,036 & 104,948 & 105,939 & 1,079,923 & 33.7\% \\
Enhancer & 768,982 & 96,123 & 96,122 & 961,227 & 30.0\% \\
Intron & 455,092 & 56,435 & 57,393 & 568,920 & 17.8\% \\
Repeat & 408,598 & 51,075 & 51,074 & 510,747 & 16.0\% \\
Promoter & 62,837 & 7,859 & 7,856 & 78,552 & 2.5\% \\
\midrule
\textbf{Total} & \textbf{2,564,545} & \textbf{316,440} & \textbf{318,384} & \textbf{3,199,369} & \textbf{100.0\%} \\
\bottomrule
\end{tabular}
}
\end{table*}

Raw genomic feature lengths exhibit massive natural variance, ranging from 1 bp to $>1.2 \times 10^6$ bp. To construct stable computational batches, sequences are dynamically cropped or padded with flanking genomic background to a fixed context window of $L=4,096$ tokens. Table~\ref{tab:feature_lengths} details the processed sequence length statistics and the \textit{Context Fit Ratio} (Fits $\le 4096$), denoting the percentage of anchors where the entire biological element fits completely within the attention window.

\begin{table*}[t]
\caption{\textbf{Processed feature lengths and context fit.} Lengths describe
the biological anchor before cropping or addition of flanking context. Most
promoters, enhancers, and exons fit within the 4,096-bp model window, whereas
many introns do not.}
\label{tab:feature_lengths}
\centering
\resizebox{0.8\textwidth}{!}{%
\begin{tabular}{@{}l r r r r r r r@{}}
\toprule
\textbf{Category} & \textbf{Min} & \textbf{P25} & \textbf{Median} & \textbf{Mean} & \textbf{P75} & \textbf{Max} & \textbf{Fits $\le 4096$} \\
\midrule
Promoter & 2,576 & 4,000 & 4,000 & 4,000 & 4,000 & 4,000 & 100.0\% \\
Enhancer & 150 & 208 & 287 & 273 & 340 & 350 & 100.0\% \\
Exon & 1 & 102 & 175 & 405 & 380 & 347,300 & 99.4\% \\
Repeat & 500 & 596 & 781 & 1,287 & 1,248 & 500,000 & 96.7\% \\
Intron & 200 & 1,002 & 2,898 & 11,335 & 9,745 & 1,240,120 & 58.0\% \\
\midrule
\textbf{Global} & \textbf{1} & \textbf{201} & \textbf{344} & \textbf{2,544} & \textbf{965} & \textbf{1,240,120} & \textbf{91.8\%} \\
\bottomrule
\end{tabular}
}
\end{table*}

These statistics show that the training windows mix anchors of very different
sizes with variable amounts of flanking sequence. They motivate evaluating
non-uniform corruption, but do not imply that local entropy directly identifies
functional sequence.

During training, real functional anchors are dynamically mixed with random genomic background regions to reinforce discriminative structural priors. Additionally, conditional sequence tags are randomly dropped 10\% of the time. Rather than explicitly utilizing this for Classifier-Free Guidance (CFG)—which we observe can cause logit saturation and structural collapse in discrete token diffusion—this unconditional dropout acts as a regularizer. It ensures the model learns a robust unconditional baseline distribution of the genome without over-indexing on structural tags. All downstream validation and test evaluations are conducted solely on true anchors.

\subsection{Sequence Formulation and Tokenization}
We employ a 1-mer tokenizer to preserve exact nucleotide resolution for ClinVar
SNV scoring and sequence reconstruction. Efficient NumPy byte mapping produces
a 28-token vocabulary containing A, C, G, T, N and specialized context tokens.
Sequences take the form
\texttt{<BOS> [HUMAN] [REGION] <DNA\_Tokens> <EOS>}, where \texttt{[REGION]}
denotes \texttt{[PROMOTER]}, \texttt{[EXON]}, \texttt{[ENHANCER]},
\texttt{[INTRON]}, \texttt{[REPEAT]}, or \texttt{[BACKGROUND]}. Training uses
random positional jitter around the anchor center; evaluation uses deterministic
center cropping. The same choice limits 4,096 tokens to approximately 4,096 bp.
Coarser k-mer or learned tokenization could increase physical coverage at a
fixed token budget, but would alter masking, decoding, and the allele-level
likelihood used here; we do not evaluate that trade-off.

\subsection{Architectural Details and Baselines}
For the controlled ClinVar comparison, we evaluate GenDA against a causal
baseline with the same depth, attention-head count, hidden dimension, and RoPE
configuration. For inpainting, we additionally report external Evo~2 and D3LM
baselines; differences in their training corpora, scale, and tokenization make
these descriptive rather than architecture-controlled comparisons.

Baseline inclusion follows two criteria: an architecture-controlled model
trained on our corpus, or an available pretrained genomic generator that can be
applied to gap completion. D3PM, MaskGIT, MDLM, and LLaDA are methodological
antecedents rather than genomic checkpoints. Other DNA diffusion and
reward-guided systems were developed for substantially shorter, specialized
regulatory-design tasks; their scalability to our 4,096-bp heterogeneous corpus
is unestablished, and reproducing them would introduce new data, conditioning,
and optimization choices. We therefore do not present their published numbers
as if they were directly comparable.

\begin{itemize}
    \item \textbf{Llama (Causal Baseline):} A standard unidirectional autoregressive model (151M parameters, 12 layers with 1024 hidden dimension).
    \item \textbf{GenDA (Ours):} A bidirectional discrete diffusion model built on a ModernBERT backbone (202M parameters, 12 layers with 1024 hidden dimension). 
\end{itemize}

The parameter difference arises primarily from the prediction heads and
feed-forward expansion ratios used by the two architectures. The comparison
therefore matches core attention scale rather than exact parameter count.

We use ModernBERT for GenDA because it natively supports bidirectional
attention over arbitrary masked positions. We initially considered converting
the Hugging Face Llama implementation to bidirectional attention, but could not
reliably verify that every causal-masking path was disabled across the model and
optimized attention kernels. Rather than risk an inadvertently causal diffusion
backbone, we use ModernBERT and retain Llama in its native causal form as the AR
baseline. Consequently, the comparison evaluates the complete bidirectional
diffusion and causal AR configurations while controlling their depth, head
count, hidden dimension, RoPE configuration, and training split.

\subsection{Training Hyperparameters and Masking}
GenDA is optimized by minimizing Cross-Entropy Loss over a maximum of 100,000 training steps with a global batch size of 128 sequences. Training follows our three-stage curriculum noise schedule. For Stages 1 and 2 (Global Scaffolding and Structural Grammar), we utilize a cosine decay learning rate schedule with a maximum learning rate of $2 \times 10^{-4}$ following a 3,000-step linear warmup. For Stage 3 (High-Resolution Refinement), the optimizer is restarted with a fresh learning rate of $1.5 \times 10^{-5}$ and a 500-step warmup to specialize on local token precision.

For density-guided masking, the local Shannon entropy scanner uses a sliding
window of $W=9$, and span lengths are sampled from a clipped Poisson
distribution with $\lambda=100$. The final GenDA configuration uses
density-guided spans for 70\% of examples and independent position-uniform
1-mer masking for 30\%. These values were selected heuristically and were not
exhaustively tuned.

\section{Reproducibility Statement}
The appendix reports the data split, model sizes, tokenization, training
schedule, masking distribution, decoding schedule, AlphaGenome contexts,
eligible-site rule, control construction, number of generated samples, and
bootstrap procedure. We separate controlled comparisons from external
references and explicitly identify heuristic choices. Code and processed
evaluation metadata will be released upon publication, subject to the licenses
of the underlying annotations and model checkpoints.

\section{Ethics and Broader Impacts}
The present study evaluates human reference-genome sequences and public
variant annotations; it does not use identifiable participant records. More
reliable genomic generation could eventually support synthetic biology and
therapeutic research, but sequence generators also create dual-use risks. Our
negative result argues against treating predictor scores as evidence that a
generated sequence is biologically valid or safe. Experimental validation,
biosafety review, and application-specific oversight remain necessary before
deployment.

\section{LLM Use Disclosure}
Large language models were used to assist with language editing, organization,
and LaTeX revision. They were not used to generate experimental measurements,
select reported results, or replace author verification of the analyses and
claims. The authors remain responsible for the manuscript's content.


\newpage
\section*{NeurIPS Paper Checklist}

\begin{enumerate}

\item {\bf Claims}
    \item[] Question: Do the main claims made in the abstract and introduction accurately reflect the paper's contributions and scope?
    \item[] Answer: \answerYes{} 
    \item[] Justification: As shown in Design premise and evaluation and results sections.

\item {\bf Limitations}
    \item[] Question: Does the paper discuss the limitations of the work performed by the authors?
    \item[] Answer: \answerYes{} 
    \item[] Justification: in section 6

\item {\bf Theory assumptions and proofs}
    \item[] Question: For each theoretical result, does the paper provide the full set of assumptions and a complete (and correct) proof?
    \item[] Answer: \answerNA{} 
    \item[] Justification: The paper does not introduce novel theoretical results or formal proofs; the contributions are primarily empirical.

    \item {\bf Experimental result reproducibility}
    \item[] Question: Does the paper fully disclose all the information needed to reproduce the main experimental results of the paper to the extent that it affects the main claims and/or conclusions of the paper (regardless of whether the code and data are provided or not)?
    \item[] Answer: \answerYes{} 
    \item[] Justification: in appendix

\item {\bf Open access to data and code}
    \item[] Question: Does the paper provide open access to the data and code, with sufficient instructions to faithfully reproduce the main experimental results, as described in supplemental material?
    \item[] Answer:  \answerNo{} 
    \item[] Justification: Code will be available upon acceptance.

\item {\bf Experimental setting/details}
    \item[] Question: Does the paper specify all the training and test details (e.g., data splits, hyperparameters, how they were chosen, type of optimizer) necessary to understand the results?
    \item[] Answer:  \answerYes{} 
    \item[] Justification: in appendix

\item {\bf Experiment statistical significance}
    \item[] Question: Does the paper report error bars suitably and correctly defined or other appropriate information about the statistical significance of the experiments?
    \item[] Answer:  \answerYes{} 
    \item[] Justification: 95\% confidence interval is included

\item {\bf Experiments compute resources}
    \item[] Question: For each experiment, does the paper provide sufficient information on the computer resources (type of compute workers, memory, time of execution) needed to reproduce the experiments?
    \item[] Answer: \answerYes{}
    \item[] Justification: We state here that all experiments were conducted on a single machine utilizing 2 NVIDA H100 GPUs, which is sufficient to reproduce the presented results.
    
\item {\bf Code of ethics}
    \item[] Question: Does the research conducted in the paper conform, in every respect, with the NeurIPS Code of Ethics \url{https://neurips.cc/public/EthicsGuidelines}?
    \item[] Answer: \answerYes{} 
    \item[] Justification: Yes, we have Ethics and Broader Impacts in the appendix

\item {\bf Broader impacts}
    \item[] Question: Does the paper discuss both potential positive societal impacts and negative societal impacts of the work performed?
    \item[] Answer: \answerYes{} 
    \item[] Justification:  Yes, we have Ethics and Broader Impacts in the appendix
    
\item {\bf Safeguards}
    \item[] Question: Does the paper describe safeguards that have been put in place for responsible release of data or models that have a high risk for misuse (e.g., pre-trained language models, image generators, or scraped datasets)?
    \item[] Answer: \answerNA{} 
    \item[] Justification: Our work utilizes existing, publicly available genomic datasets. The specific models developed in this study serve as academic proofs-of-concept and do not possess capabilities that pose an immediate, high risk for misuse requiring gated release.

\item {\bf Licenses for existing assets}
    \item[] Question: Are the creators or original owners of assets (e.g., code, data, models), used in the paper, properly credited and are the license and terms of use explicitly mentioned and properly respected?
    \item[] Answer: \answerYes{} 
    \item[] Justification: We are the original owners of the model. The dataset is public.

\item {\bf New assets}
    \item[] Question: Are new assets introduced in the paper well documented and is the documentation provided alongside the assets?
    \item[] Answer: \answerNA{} 
    \item[] Justification: The paper relies on existing, publicly available datasets and does not introduce new standalone assets.

\item {\bf Crowdsourcing and research with human subjects}
    \item[] Question: For crowdsourcing experiments and research with human subjects, does the paper include the full text of instructions given to participants and screenshots, if applicable, as well as details about compensation (if any)? 
    \item[] Answer:  \answerNA{} 
    \item[] Justification: The research does not involve crowdsourcing or human subjects.

\item {\bf Institutional review board (IRB) approvals or equivalent for research with human subjects}
    \item[] Question: Does the paper describe potential risks incurred by study participants, whether such risks were disclosed to the subjects, and whether Institutional Review Board (IRB) approvals (or an equivalent approval/review based on the requirements of your country or institution) were obtained?
    \item[] Answer: \answerNA{} 
    \item[] Justification: The research does not involve human subjects.

\item {\bf Declaration of LLM usage}
    \item[] Question: Does the paper describe the usage of LLMs if it is an important, original, or non-standard component of the core methods in this research? Note that if the LLM is used only for writing, editing, or formatting purposes and does \emph{not} impact the core methodology, scientific rigor, or originality of the research, declaration is not required.
    \item[] Answer: \answerNA{} 
    \item[] Justification: We did not use external LLMs as a methodological component. While our core contribution is a DNA language model, its architecture and training pipeline are fully described in the main text as our primary method.

\end{enumerate}

\end{document}